\documentclass[a4paper, conference]{IEEEtran}
\usepackage{graphicx}

\usepackage{algorithm}
\usepackage{algorithmic}

\usepackage[numbers,sort&compress]{natbib}

\begin{document}

\title{Characterizing Lifting and Lowering Activities with Insole FSR sensors in Industrial Exoskeletons
}





	
\author{\IEEEauthorblockN{Luis A. Mateos, Jes\'{u}s Ortiz, Stefano Toxiri, \\Jorge Fern\'{a}ndez, Jawad Masood, Darwin G. Caldwell}
	\IEEEauthorblockA{Department of Advanced Robotics\\
		Istituto Italiano di Tecnologia, Via Morego, 30, 16163 Genova, Italy\\
		Email: lamateos@mit.edu | jesus.ortiz; stefano.toxiri; jorge.fernandez; jawad.masood; darwin.caldwell \{@iit.it \} }
	
}


%



\maketitle
\thispagestyle{empty}
\pagestyle{empty}


\begin{abstract}
	
	
	This paper presents an insole FSR (Force Sensing Resistor) to dynamically detect weight variations in an exoskeleton system. 
	The proposed methodology is intended for tasks of lifting and lowering heavy weights with an industrial exoskeleton to substantially reduce spinal loads during these manual handling activities. 
	
	Instead of extensively placing high-dense force sensors by covering the whole plantar area, as most of commercial applications do, we integrate only a few force sensors in specific plantar area, so that the sensory system is not restricted to an individual foot size and shape, and on the other hand has relatively lower material cost. 
	
	Industrial exoskeletons are intended to assist workers when handling heavy goods. With this in mind, wearers are not able to use their hands to control the exoskeleton since they use them to handle the goods. Therefore, the exoskeleton controller is required to indirectly infer how much and when the wearer requires assistance for lifting or lowering a heavy weight. 
	Our approach of dynamically detect and characterize the increment/decrement of weight, as well as the rising/falling edge, enables the exoskeleton's controller to trigger the request of assistive force to the actuators.
	


\end{abstract}


\section{Introduction}
\label{sec:introduction}


It is a common practice in laboratory that the exoskeleton's wearer controls the system by indirect measures, such as electromyography (EMG) \cite{1244649, 6798669} or force/torque measurement generated from interaction \cite{4399147}\cite{4428426}. 
This means that the wearer does not directly control the exoskeleton with a joystick as common cases for robots or mechanical systems. The reason behind is that the hands and in some cases the full body of the wearer are expected to be free. 
In this context, the ideal exoskeleton should be able to infer the intention of the user by indirect and reliable measures in which the user does not need to pay attention to control it, instead the wearer moves naturally as the system is able to follow the wearer movements like a 'shadow'.

EMG signals contain rich information that can be used to control and drive an exoskeleton robot and similar man-machine devices in both rehabilitation and assistance \cite{5420670}. 
The wearable robot Hybrid Assistive Limb (HAL) developed by Sankai et al. \cite{1244649}, is comprised of four actuators for hip and knee assistance and uses feedback from both EMG and force sensors. Similarly, Li et al. \cite{6655907} \cite{6798669} and Fleischer et al. \cite{4209461} \cite{4560058} use EMG signals to control an exoskeleton.

\begin{figure}[thpb]
	\centering
		\includegraphics[width=0.42\textwidth]{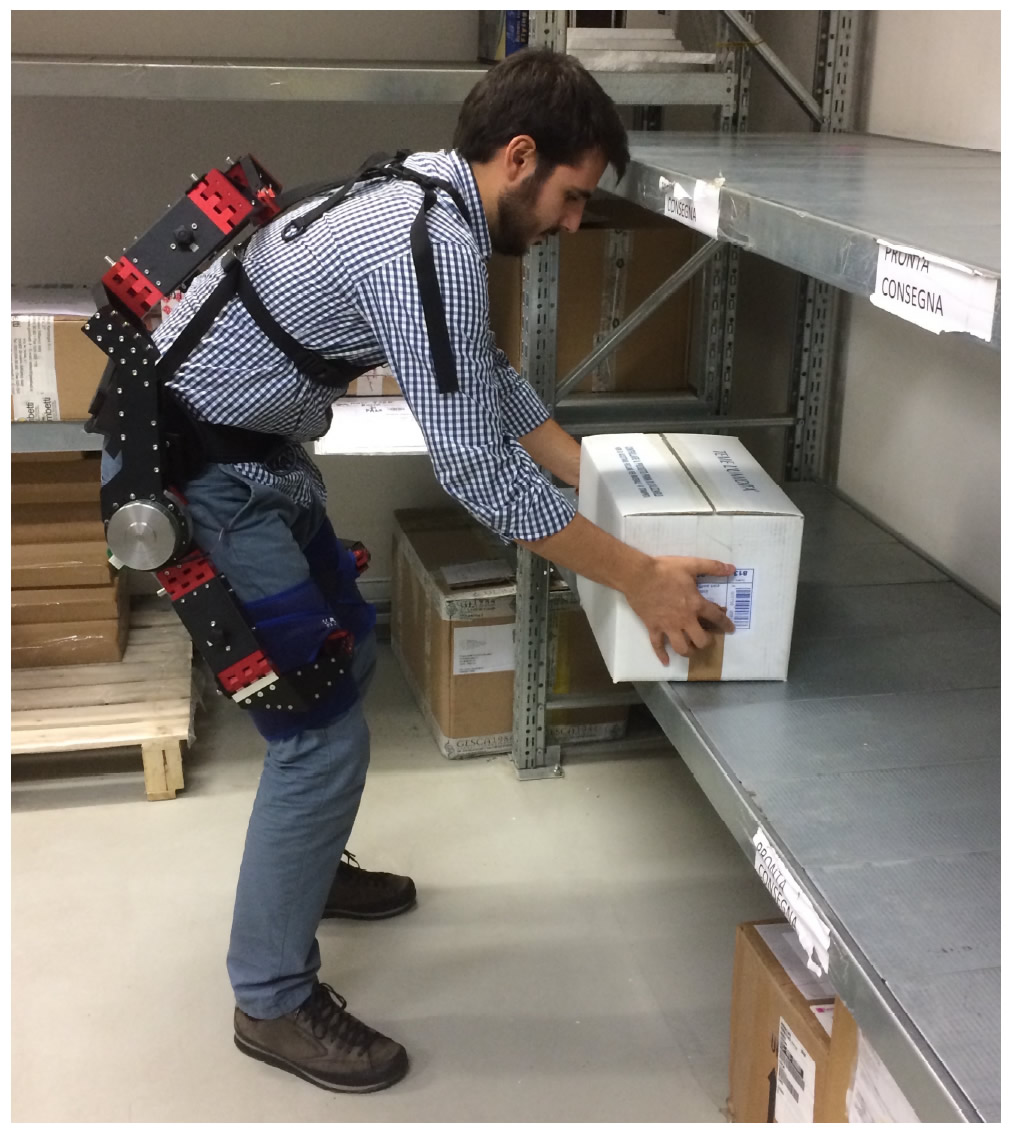}
		\caption{Industrial exoskeleton Robo-mate. }
		\label{ffig1}
\end{figure}

However, in contrast to laboratory environments, where the temperature and humidity are controlled in an indoor setting, industrial environments present a big challenge for monitoring technologies that are glued or fasten to the operator's body.

The robotic assistance of an industrial exoskeleton may alternatively be regulated according to external conditions, such as the weight of the object to be handled \cite{7523605}. In this case, force sensors are integrated in the wearer's footwear or within the plantar area of the exoskeleton where the wearer is standing on. Hence, from the force sensors it is possible to detect the variation of planar pressure which enables the exoskeleton's controller to trigger the assistive force.

\begin{figure*}[thpb]
	\centering
		\includegraphics[width=\textwidth]{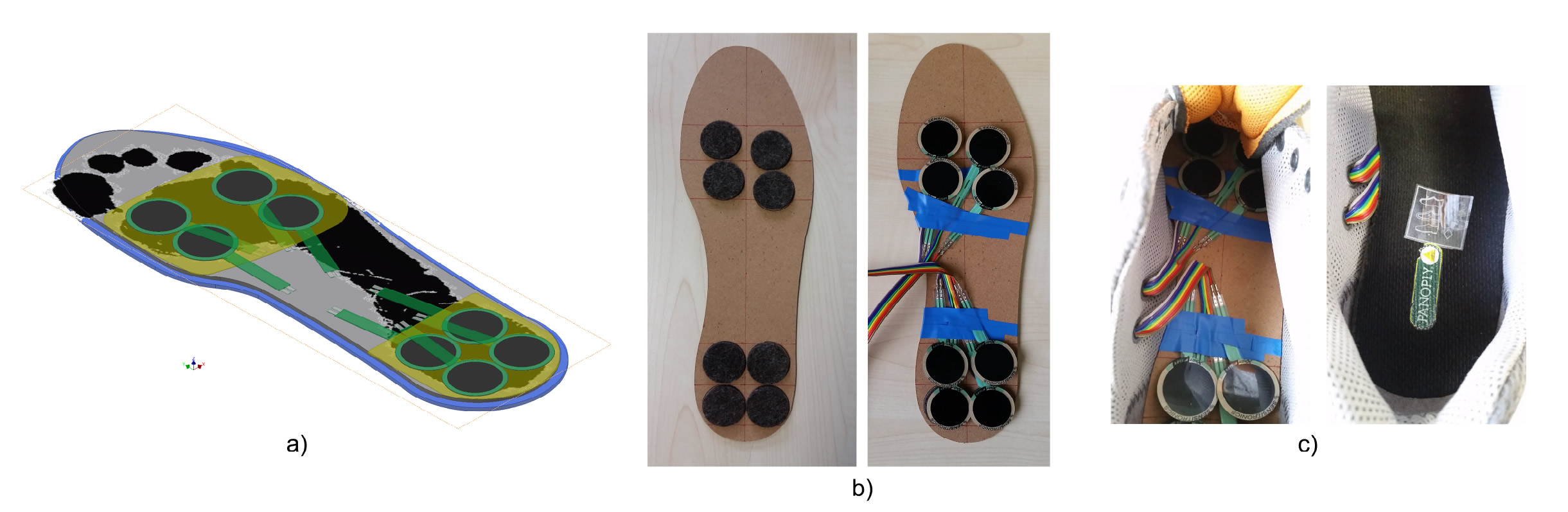}
		\caption{Insole FSR system module. a) Sensor arrangement under foot, 8 force sensors divided in two modules (metatarsals  on the $front$ and heel on the $back$). b) Insole module with force sensors.  c) Insole integrated in the shoe.}
		\label{figsensorresis}
\end{figure*}

The plantar pressure systems available on the market, such as Ekso Bionics  \cite{EKSO1}, Pedar \cite{PEDAR1} and F-Scan \cite{EKSO2}, embed high dense of force sensors to cover the complete plantar area in order to overcome the acceleration force.  Due to the extensive coverage of the sensors, these systems are highly customized to the individual foot shape and size, priced over 15,000USD each. Thus, the applications purely depending on pressure measurement are limited to laboratory usage due to their high cost. 

We intend to develop a sensory system, which is portable in common footwear and has low cost in material, so that it can be applied in industrial scenarios. We reduce the coverage of the sensors to only the heel  and metatarsals area, where most of the foot pressure would be detected during upright standing position.

Our endeavor represents part of the Robo-Mate project \cite{robomate}, which aims to deliver a powerful exoskeleton, easy and intuitive to command. The goal of Robo-Mate is to develop a wearable powered device to substantially reduce spinal loads during lifting activities.

This paper is structured as follows: Section \ref{sec:forcesensors} describes the properties of common thin film force sensors and their applications in exoskeleton-foot sensors. 
Section \ref{sec:design} presents the concept and hardware prototype of the insole system with force sensors FSR to be integrated in the exoskeleton system. 
Section \ref{sec:calibequa} describes the circuit design for the insole FSR sensors. 
Section \ref{sec:experiments} presents the experimental results from the FSR and the algorithm to detect and characterize  the lifting and lowering activities. 
Conclusions are proposed in Section \ref{sec:conclusion}, which also points at some of the next steps and future challenges expected in the future work.



\section{Force sensors}
\label{sec:forcesensors}

%

Initial attempts to have foot-force sensors to measure the pressure applied in the front and rear parts of the foot (ball and heel of foot) are described in \cite{1244649}. Their proposed floor reaction force (FRF) sensor detects air pressure changes sensed when foot-pressure is applied to their customized air-pressure force sensors. Up to today the electrical force sensors instead of air pressure based sensors are widely applied. For instance, thin film sensors such as flexiforce (Tekscan\textregistered) or shunt force sensor (Sensitronics\textregistered) are ideal for integration in footwear, as they have small thickness of less than one millimeter to fit into the limited space of the footwear. In addition, these types of force sensors are an inexpensive and lightweight alternative to multi-axis force/torque sensors.

These types of force sensors have been integrated in biped robots for gait investigation\cite{6798963}, where they are placed on the bottom of the robot's feet to provide feedback for the control system. 
Also, force sensors have been integrated inside shoes to monitor the foot pressure of a person when walking and running \cite{5695791}. 
Moreover, in-pipe robots integrated these types of thin-film force sensors. For instance, the DeWaLoP in-pipe robot integrates six force sensors on each of its wheeled-legs to measure the force exerted when the robot extend the legs to become a rigid structure inside the pipe \cite{6766548}.

A force sensor can be modeled by Equation \ref{eqvirtualforcesensor2}

\begin{equation}
	f(x) =  \left(\frac{1}{Rx}\right)
	\label{eqvirtualforcesensor2}
\end{equation}

\noindent where $x$ presents the received force by the sensor and $f(x)$ presents the generated resistance based on given force,  a constant $R$ is to model a smooth curve response. 



Commercial force cells are commonly thick  with heights of around $11mm$ and expensive, if compared to 6USD thin film force sensors with height of $0.1"$ ($0.254mm$). In our design we considered thin film force sensors, since these can be integrated under the shoe  with the minimal height increment. 

Commercial thin film force sensors have similar mechanical and electrical characteristics, which were analyzed in \cite{6798963}\cite{5944336} and \cite{Hall20083492}. For our specific requirements, to dynamically detect the weight of a person when lifting or lowering a heavy object, we analyzed a couple of the most used commercial sensors: flexiforce and shunt force sensor. For our application we selected the shunt force sensor \cite{Sensitronics}. 
The key element for our selection is that the shunt force sensor provides relatively higher resolution in our use cases, from 0 to $98N$ ($10kg$), the sensor range is from 0 to $441N$ ($45kg$).


\section{Design}
\label{sec:design}


\begin{figure}[b]
	\centering
		\includegraphics[width=0.4\textwidth]{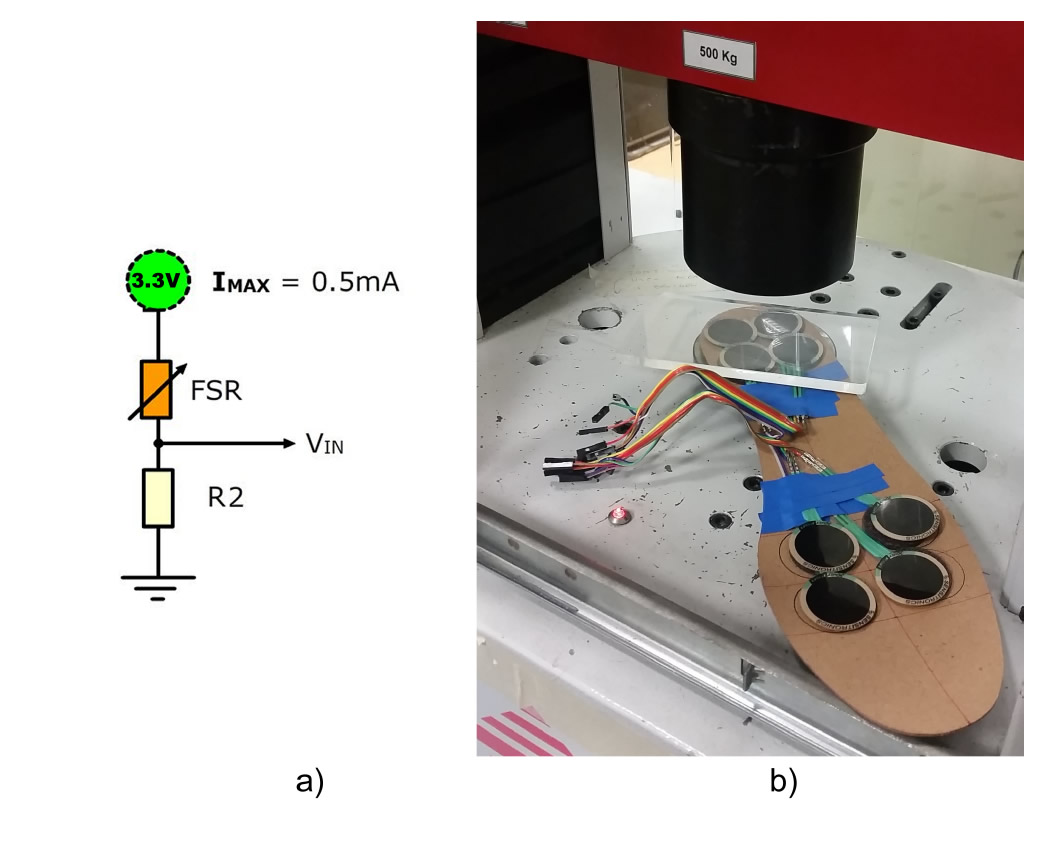}
		\caption{a) Protection circuit for FSR sensors. b) Compression test machine for FSR calibration to responses ($0$ to $441N$). }
		\label{figforcesensorcalib}
\end{figure}

The intended sensorised insole system is integrated to the operator's shoes which is able to dynamically measure weights and pressure changes  when a person is lifting or lowering a heavy weight.  
The aim of the FSR insole system is to indirectly infer when the exoskeleton should assist the wearer and also to detect how much assistance is required. 

To clearly differentiate the states of the operator, the force sensors located in the insole of the operator are required to have a good resolution, while physically being as thin as possible for not increasing the height of the operator.



The proposed module design integrates eight force sensors per foot. They are divided into two groups of four sensors, located in the peak pressure points of a normal foot: the metatarsals and the heel \cite{6347549}. In this configuration, the insole FSR system have a force resolution ranging from $0$ to $720kg$ if all the insole FSR sensors are at their maximum capacity. However, we are only considering a maximum of $98N$ ($10kg$) per sensor, resulting in a maximum of $1568N$ ($160kg$).

Each insole module consists of an inner base where the sensors are attached. The base is covered by the insole of the shoe to protect the sensors, see figure \ref{figsensorresis}.


\section{Force sensors calibration and equalization}
\label{sec:calibequa}


Commercial thin film force sensors FSR have several drawbacks that make them not suitable for precise measurement of force. Their resistance response respect to the force is nonlinear and from sensor to sensor is different. This means that each sensor will respond with different resistance values even when the applied force is the same. 
Therefore, each thin film FSR sensor requires to be calibrated. 

\subsection{Force Sensor Calibration} 

The FSR sensors integrate a conductive ink which have a maximum current limit. Figure \ref{figforcesensorcalib}$a$ shows the circuit to protect and obtain the force-resistance curves from a FSR sensor. 

The circuit consists of a voltage divisors. It is a current limiter to $0.5mA$, which is the maximum current supported by the sensor \cite{shunt}, and it is also set to read the voltage at the FSR sensor. In this configuration, $R_2$ is calculated from the FSR resistance values, averaging the higher and lower limits to obtain a considerable ratio to differentiate between the levels, see Table \ref{tbl:fsrtable}.

\begin{table}[h]
\centering
\caption{Voltage divisor values for FSR.}
\begin{tabular}{|l|l|l|l|l|}
\hline
Force level & R1  & R2  & Ratio  & Vout \\ 
(in N) & (FSR sensor) & (Fixed) & R2/(R1+R2) & (in V) \\ \hline
$490N$ (MAX)  & $100\Omega$             & $5.6k\Omega$      & 0.85               & 2.80V   \\ \hline
$200N$        & $3000\Omega$            & $5.6k\Omega$      & 0.65               & 2.15V   \\ \hline
$40N$ (min)   & $10000\Omega$            & $5.6k\Omega$      & 0.36              & 1.18V   \\ \hline
\end{tabular}

\label{tbl:fsrtable}
\end{table}

In order to calibrate the FSR sensors we used a compression test machine, see figure \ref{figforcesensorcalib}$b$. We applied forces to each sensor in the range of 0 to $441N$ ($45kg$).  We loaded the FSR sensor, applying the force in steps of $9.8N$ ($1kg$) and holding for a couple of seconds, then release the force. 

\begin{figure}[b]
	\centering
		\includegraphics[width=0.4\textwidth]{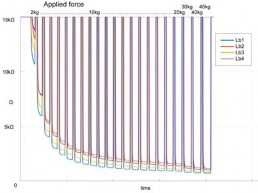}
		\caption{ Force-resistance curves from the $Left$ foot $back$ FSR sensors, revealing different responses to the same applied force. }
		\label{curves}
\end{figure}

The voltage readings from the FSR sensor $V_{FSR}$ are converted to digital levels with an ADC (1024 steps of resolution). The voltage input $V_{IN} = 3.3V $, and the steps are calculated $Steps_{FSR} = V_{FSR} \times 1024 / V_{IN}$.

The resistance from a FSR sensor $R_{FSR} $ is obtained by Equation \ref{eq2}, where $R_2$ is the fix resistor in a voltage divisor configuration.

\begin{equation}
	R_{FSR} =  \left(\frac{V_{IN} R_2}{V_{IN} (Steps_{FSR}/ADC)}\right) - R2
	\label{eq2}
\end{equation}

Once the sensors are calibrated, their force-resistance curves are know. These curves reveals that the sensors are considerably different, as shown in figure  \ref{curves}.

\begin{figure*}[thpb]
	\centering
		\includegraphics[width=0.95\textwidth]{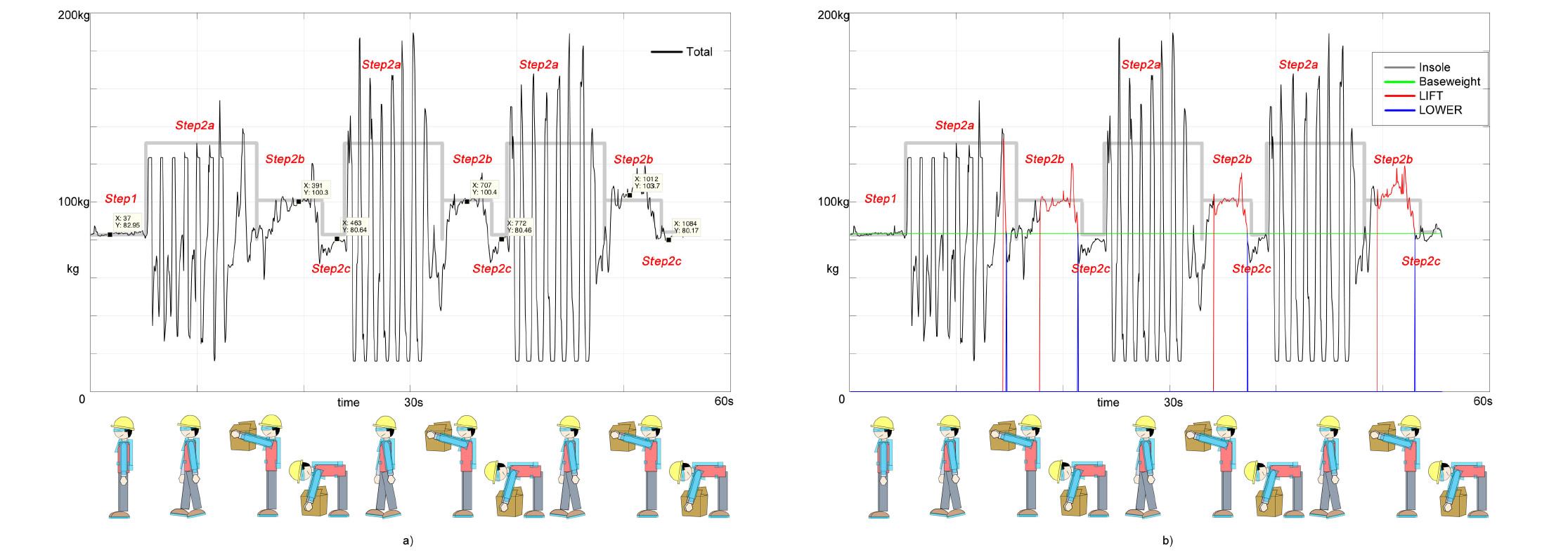}
		\caption{Experiment I.  a) Weight detection. b) Lifting and lowering detection.}
		\label{e14}
\end{figure*}





\subsection{Sensor Configurations}
\label{senconf}

The proposed insole design integrates eight force sensors per foot, making in total 16 force sensors. This means that ideally 16 input channels are required to monitor and process the sensors data from the feet. 

\textbf{Note:} 
The FSR sensor cannot be group together in modules since they have different force-resistance curves. Even if the sensors are equalized, their responses are still different. Thus, the grouping of sensors will lead to erroneous results and we must take the value of each FSR sensor independently.





\section{Experiments}
\label{sec:experiments}



The objective of the experiments is to register the weight detected by the FSR sensors when all sensors are added independently, and also to detect and characterize the lifting and lowering activities.


\subsection{Experimental Setting}

The experimental setting consists of four micro-controller boards, four radio transceiver and a computer to store and analyze the gathered data. 
In this setting, two Personal Area Networks (PAN) are created, one per foot. The aim of this setting is to enable the subject to move and walk naturally without cables between the computer and the insole FSR when performing the experiments.  


\subsection{Experiment I}

The experiments were performed in the insole configuration with 8 FSR sensors per foot.

\textbf{Experiment I steps:}
\begin{enumerate}
	\item Sensors are initialized (the person's weight ($83kg$) is registered as $base$ weight).
	\item Repeat 3 times: \\
	a) Walk few steps.\\
	b) Lift up $18.6kg$ weight from the ground.\\
	c) Lower $18.6kg$ weight from the ground.
\end{enumerate}

\textbf{Experiment II steps:}
\begin{enumerate}
	\item Sensors are initialized (the person's weight ($83kg$) is registered as $base$ weight).
	\item Walk few steps.
	\item Repeat 3 times: \\
	a) Lift up $9.3kg$ weight from the ground.\\
	b) Lower $9.3kg$ weight from the ground.
\end{enumerate}

\begin{figure*}[thpb]
	\centering
		\includegraphics[width=0.95\textwidth]{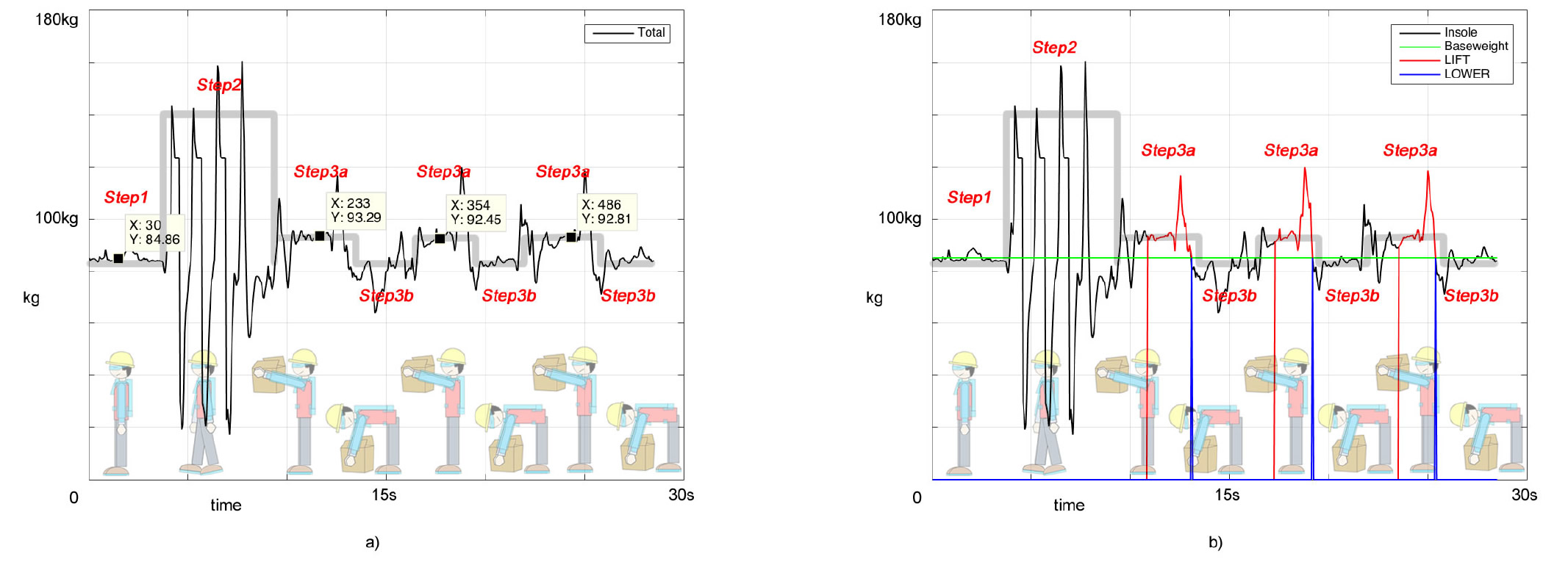}
		\caption{Experiment II. a) Weight detection. b) Lifting and lowering detection.}
		\label{e13}
\end{figure*}






\subsection{Weight Detection Results}

In Experiment I (see figure \ref{e14}), the results show a correlation of the real applied force to the force read by the sensors. The $base$ weight of the subject is $83kg$ and the insole FSR sensors detected $82.95kg$. When lifting the heavy object of $18.6kg$ the insole detected the weight $\pm15\%$. Also, when lowering the heavy weight during the repetitions, the insole system reads a similar $base$ weight $\pm5\%$

In Experiment II (see figure \ref{e13}), the results are similar to Experiment I, the results show that the weight detected is consistent with the ground truth varying $\pm20\%$.

\begin{table}[]
\centering
\caption{Experiment I insole FSR sensors values.}
\label{my-label1}
\begin{tabular}{|l|l|l|l|}
\hline
Independent     & Repetition 1 & Repetition 2 & Repetition 3 \\
configuration     & (in $kg$) & (in $kg$) & (in $kg$) \\ \hline \hline

$base$ weight     & $82.95kg$              &   $82.95kg$                   &    $82.95kg$                  \\ \hline
lowering $18.6kg$ & $80.64kg$              & $80.46kg$                & $80.17kg$                \\ \hline
$error$ percentage &   $ -2.78\%$           &    $-3\%$               &    $-3.35\%$               \\ \hline \hline

lifting $18.6kg$  & $100.3kg$              & $100.4kg$              & $103.7kg$              \\ \hline
weight detected $-$ &                      &                       &                       \\ 
$base$ weight &   $17.35kg$                   &  $17.45kg$                    &   $20.75kg$                   \\ \hline
$error$ percentage &   $ -6.72\%$           &    $-6.18\%$               &    $+11.56\%$               \\ \hline
\end{tabular}

\end{table}

\begin{table}[]
\centering
\caption{Experiment II insole FSR sensors values.}
\label{my-label1}
\begin{tabular}{|l|l|l|l|}
\hline
Independent     & Repetition 1 & Repetition 2 & Repetition 3 \\
configuration     & (in $kg$) & (in $kg$) & (in $kg$) \\ \hline \hline

$base$ weight     & $84.86kg$              &   $84.86kg$                   &    $84.86kg$                  \\ \hline

lifting $9.3kg$  & $93.29kg$              & $92.45kg$              & $92.81kg$              \\ \hline
weight detected $-$ &                      &                       &                       \\ 
$base$ weight &   $8.43kg$                   &  $7.59kg$                    &   $7.95kg$                   \\ \hline
$error$ percentage &   $ -9.35\%$           &    $-18.39\%$               &    $-14.5\%$               \\ \hline
\end{tabular}

\end{table}


\subsection{Detection of Lifting and Lowering Activities }

The detection of lifting and lowering are performed dynamically, this means that the sensors values require a quarter of a second to process and detect weight. In this context, the algorithm is based on weight detection over a time.

The lifting Algorithm \ref{a1}, requires to know the user's weight so it can be compared to the data read by the insole FSR sensors. From the experimental data, see figures \ref{e14} and \ref{e13}, we know that the detected weight of the person goes lower its $base$ weight before a lifting. This is due to the dynamics of the bending movement of the person to pick an object from the ground. Thus, our algorithm sets a flag to prepare the lifting detection when this happens.
Once the $prepare$ $for$ $lifting$ flag is set, the algorithm detects increments of weight over a time frame, in our case is one quarter of a second, to set the state of $LIFTING$. 

The lowering Algorithm \ref{a2}, also requires to know the user's weight so it can be compared to the data read by the insole FSR sensors. In the same way as the lifting detection, from the experimental data we know that before lowering the detected weight goes higher than the $base$ weight plus the lifted weight. This is due to the dynamics of the movements when the person is positioning to lower the heavy weight. In this way, our algorithm sets a flag to prepare the lowering detection.
When the $prepare$ $for$ $lowering$ flag is set, the algorithm detects the decrement of weight and when it goes lower the $base$ weight then the algorithm set the state of $LOWERING$.

\textbf{Note:}
In Experiment I, see figure \ref{e14}, the results show four lifting and lowering actions instead of three. However these detections are correct. At the beginning of the experiment, two packs of $6\times1.5L$ water bottles are pile one over the other and the user carries one pack with the right arm to put it on his right side to eventually carry one pack on each hand. In this link we collected the videos from these experiments\footnote{http://watsonjosh2.wix.com/insolefsr}.

In Experiment II, see figure \ref{e13}, the results clearly show the lifting and lowering of the heavy object.

\begin{algorithm}
\caption{Detect LIFTING }
\label{a1}
\begin{algorithmic} 

\REQUIRE $base_{weight}$ 
\ENSURE user is standing still for 5 seconds
\STATE $base_{weight} \leftarrow User_{weight}$

\IF {$detected_{weight} <  base_{weight}$}
\STATE PRE LIFTING
\ENDIF

\IF{$PRE LIFTING$ \\
\AND {$detected_{weight} >  base_{weight}$ } \\
\AND {$detection_{time} >  time_{threshold}$}}
\STATE LIFTING
\ENDIF

\end{algorithmic}
\end{algorithm}

\begin{algorithm}
\caption{Detect LOWERING }
\label{a2}
\begin{algorithmic} 

\REQUIRE $base_{weight}$ 
\ENSURE current state LIFTING
\IF{$LIFTING$ \\
\AND {$detected_{weight} >  base_{weight} + lifted_{weight}$}}
\STATE PRE LOWERING
\ENDIF

\IF{$PRE LOWERING$ \\
\AND {$detected_{weight} <  base_{weight}$}}
\STATE LOWERING
\ENDIF

\end{algorithmic}
\end{algorithm}










\section{Conclusion and future work}
\label{sec:conclusion}


The results show that the proposed insole FSR system is able to detect the weight of a person and the lifted weight, by integrating force sensors located under the shoes of a person. 
The FSR insole system integrate 8 force sensors per foot, divided in two modules, one module located under the heel area and the other under the foot metatarsals. In this configuration, the FSR insole system is able to measure the forces between $0$ to $160kg$. Nevertheless, it can measure a maximum of $720kg$.

We computed the insole FSR sensors independently, since each sensor have a unique response and force-resistance curve. The results from adding the detected weight from all sensor are consistent to the ground truth with an average variation of $\pm15\%$.



Our proposed sensory system is portable to common foot wears or exoskeletons and nonrestrictive to individual foot due to its modularized sensor configuration instead of the classic configuration of one-piece extensively plantar coverage. On the other hand the FSR insole system has relatively low cost in production due to the minimum number of force sensors being integrated, so that it has the potential to be applied in industrial use.





\addtolength{\textheight}{-3cm}   

\section*{Acknowledgment}

The authors acknowledge the contributions of the entire Robo-Mate research consortium towards the creation of the Robo-Mate exoskeleton concept.

The research leading to these results has received funding from the European Union$'$s Seventh Framework Programme for research, technological development and demonstration under grant agreement n$^\circ$ 608979, and from the People Programme (Marie Curie Actions) of the European Union$'$s Seventh Framework Programme FP7/2007-2013/ for research, technological development and demonstration under REA grant agreement n$^\circ$ 608022.


\bibliographystyle{plain}
\bibliography{references}

\end{document}